\documentclass{article}

\usepackage[final]{neurips_2018}

\usepackage[utf8]{inputenc} 
\usepackage[T1]{fontenc}    
\usepackage{hyperref}       
\usepackage{url}            
\usepackage{booktabs}       
\usepackage{amsfonts}       
\usepackage{nicefrac}       
\usepackage{microtype}      
\usepackage{graphicx}
\usepackage{wrapfig}

\title{Rethinking Layer-wise Feature Amounts in Convolutional Neural Network Architectures}

\author{
  Martin Mundt \\
  Goethe University \\
  Frankfurt Institute for Advanced Studies \\
  \texttt{mmundt@em.uni-frankfurt.de}
  \And
  Sagnik Majumder \\
  Goethe University\\
  \texttt{majumder@ccc.cs.uni-frankfurt.de} \\
  \AND
  Tobias Weis \\
  Goethe University\\
  \texttt{weis@ccc.cs.uni-frankfurt.de}
  \And
  Visvanathan Ramesh \\
  Goethe University\\
  Frankfurt Institute for Advanced Studies \\
  \texttt{vramesh@em.uni-frankfurt.de}
}

\begin{document}

\maketitle

\begin{abstract}
We characterize convolutional neural networks with respect to the relative amount of features per layer. Using a skew normal distribution as a parametrized framework, we investigate the common assumption of monotonously increasing feature-counts with higher layers of architecture designs. Our evaluation on models with VGG-type layers on the MNIST, Fashion-MNIST and CIFAR-10 image classification benchmarks provides evidence that motivates rethinking of our common assumption: architectures that favor larger early layers seem to yield better accuracy.
\end{abstract}

\section{Introduction and motivation}
Deep learning practices that are empirically confirmed to be valuable often turn into rules of thumb to be used by the community. One such rule of thumb is the historically grown custom of increasing the number of features (for convolutions synonymous with kernel or filter) with increasing depth of a convolutional neural network (CNN). Perpetuated by perhaps the simplicity and large success of the VGG architecture \cite{Simonyan2015}, more recent work such as residual networks \cite{He2016} or densely connected networks \cite{Huang2016} still follow this design principle. While such works achieve progress through modifying connectivity structure, changing the task or depth of the network, we, the machine learning community, tend to leave the principle of stacking $3 \times 3$ convolutions with monotonously increasing feature amounts per layer untouched. For other advances in tasks such as semantic image segmentation \cite{Ronneberger2015, Badrinarayanan2017}, the encoder strictly follows this pattern and on top mirrors the pattern in the decoder. Even though some work, such as the "network in network" architecture \cite{Lin2013}, deviates and explores alternatives in design, many architectures \cite{Ronneberger2015, Badrinarayanan2017, Huang2016, He2016} seem to inherit the simple VGG-style of keeping or doubling the amount of features from one layer to another.  
Apart from the empirically demonstrated effectiveness, a core assumption can be hypothesized as follows: lower layers of CNNs learn more primitive features whereas higher layers learn more abstract features. Thus, our assumption could be to increase the amount of learnable features in higher layers to in turn provide enough representational capacity for a rich encoding.

In this work we propose a simple three-parameter univariate skew normal distribution to parametrize a family of neural networks. By changing the distribution's parameters, we shift a constant amount of features and map them to architectures with monotonously decreasing, increasing and normally distributed feature amounts per layer. While the exact choice of distribution is of empirical nature, a three-dimensional mathematical description allows for an intuitive model characterization. We train 200 model variants by conducting a grid-search on the distribution's parameters on three popular image classification datasets: MNIST \cite{LeCun1998}, CIFAR-10 \cite{Krizhevsky2009}, Fashion-MNIST \cite{Xiao2017}. We show that the commonly picked subset of monotonously increasing feature amounts per layer seems to be outperformed in terms of accuracy by architectures that favor larger early layers. 
We hope to inspire to rethink our CNN design intuition and to stimulate further analysis for future models. 

\section{Parametrizing distribution of features across layers}
For the purpose of parametrization and characterization of common and uncommon CNN design, we have chosen the probability density function (PDF) of the univariate three parameter skew normal distribution. We use three parameters in order to be able to generate curves with varying location of the maximum peak with different sharpness, as defined by the location (mean) $\xi$ and scale (variance) $\omega$  respectively. We also require the shape (skew) $\alpha$ to adjust the slope in positive or negative direction. This results in the following PDF:
\begin{equation}
	\frac{1}{2 \pi} \frac{2}{\omega \sqrt{2 \pi}} e^{- \frac{(x - \xi)^{2}}{2 \omega ^{2}}} \frac{1}{2} [1 + erf(\frac{\alpha \frac{x - \xi}{\omega}}{\sqrt{2}}) ]
	\label{eq:skewnormal}
\end{equation}

\begin{figure}[t]
	\includegraphics[width=\textwidth]{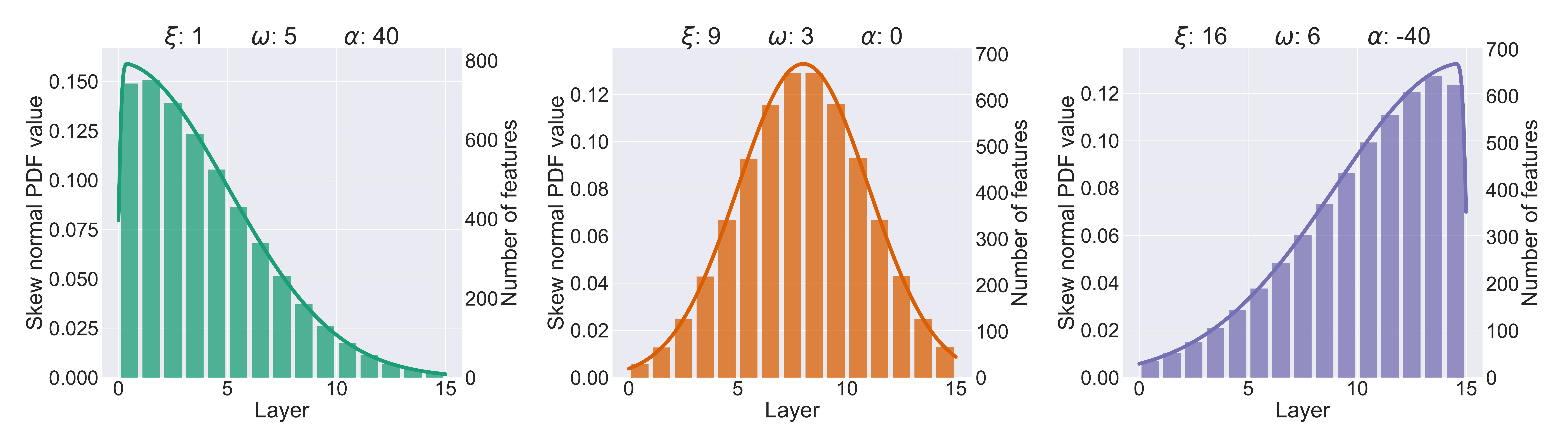}
	\caption{Three examples of skew normal PDFs (solid line), integrated layer bins and mapped amount of features. All three architectures have the same amount of overall features. Architectures with PDF parameters as depicted in the right panel most resemble traditional CNN designs. Architectures parametrized by the mid and specifically the left panel are not commonly found in the literature.}
	\label{fig:distributions}
\end{figure}

Here, $erf(x)$ is the error function. To apply this PDF to CNNs, we specify the number of layers and overall features, e.g. the total number of features in a 16 layered $3 \times 3$ convolutional VGG-D architecture. We then use trapezoidal integration to calculate one integrated value per layer. The resulting discretized distribution is scaled by the overall number of features. Using this process we can generate a family of architectures while keeping the number of layers and overall amount of features constant. We visualize three examples of the PDF, the integrated discretized layer bins, as well as the amount of features per layer in figure \ref{fig:distributions}. The figure shows three examples, corresponding to architectures with maximum amount of features in the first, middle and last layers. The latter, depicted in the right panel, is an example that is similar to the original design of the VGG-D architecture.

\section{Characterization of VGG filter distributions}
\begin{figure}[t]
	\includegraphics[width=\textwidth]{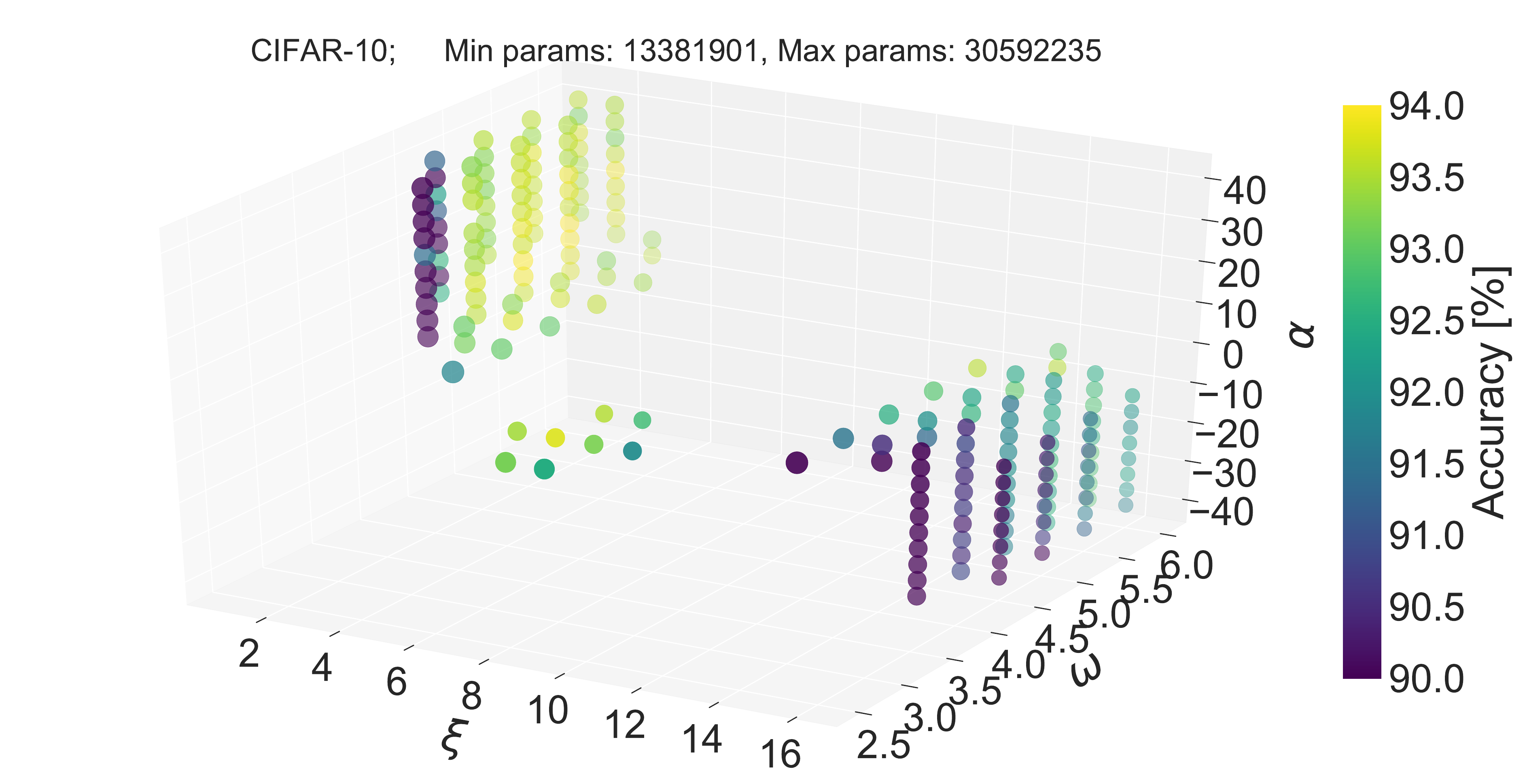} \\
	\includegraphics[width=0.5 \textwidth]{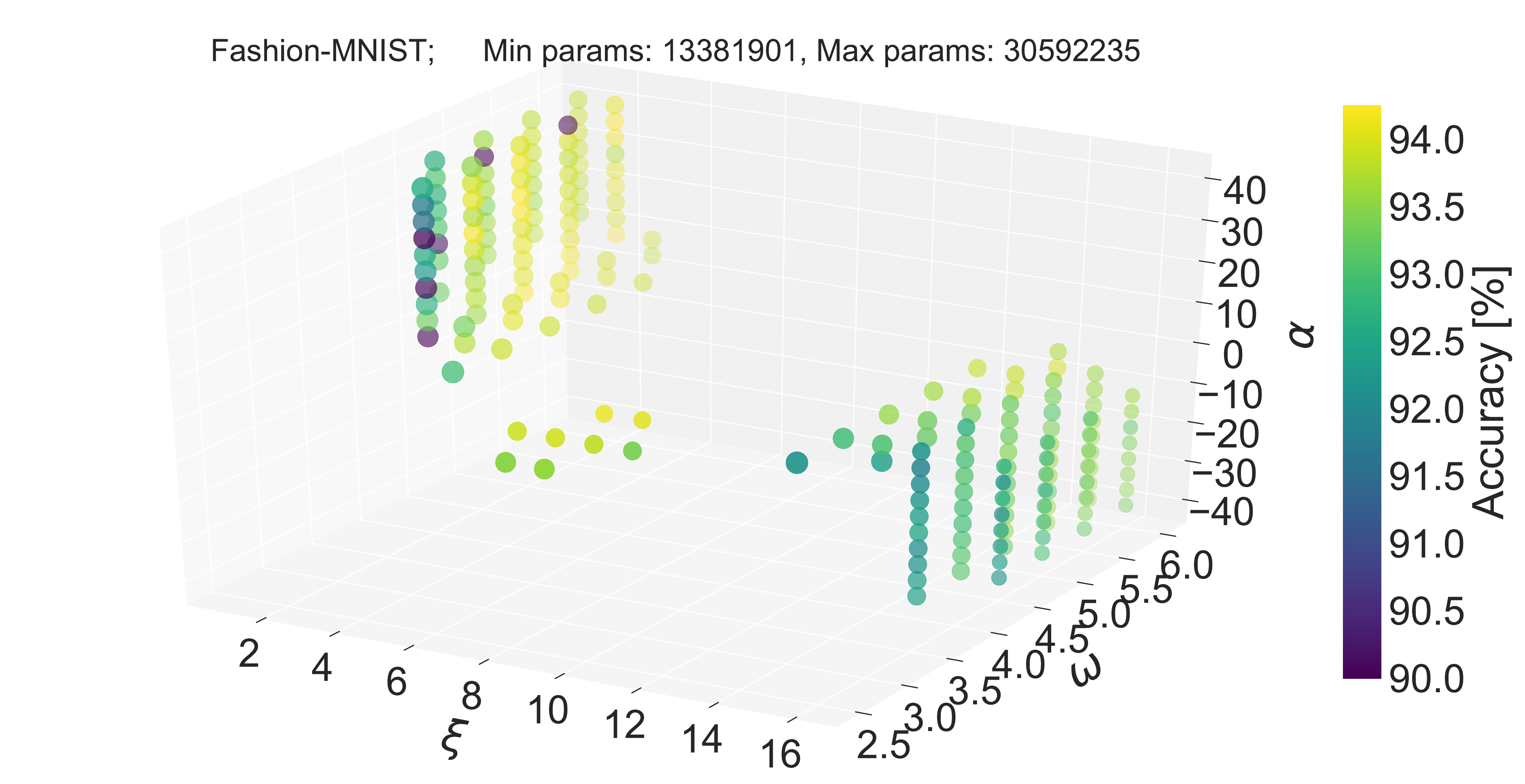}
	\includegraphics[width=0.5\textwidth]{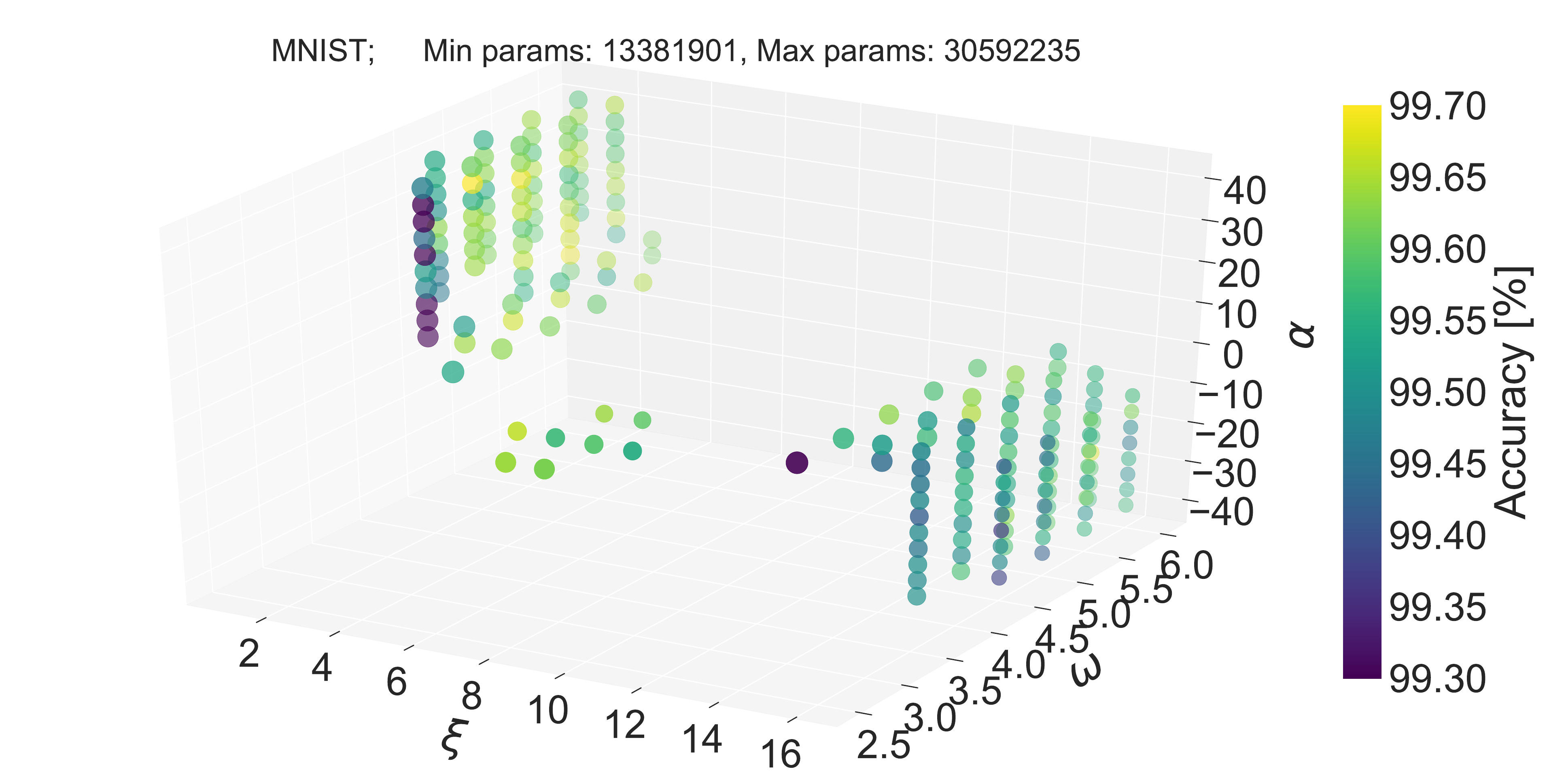}
	\caption{Validation accuracy (in color) of 16-layer VGG-type architectures parametrized through combinations of parameters $\xi, \omega, \alpha$ of a univariate skew normal distribution. While all models have approximately the same amount of features, total parameter amounts can vary as indicated by marker size. The accuracy range has been cut-off at the bottom for better visual perception.}
	\label{fig:results}
\end{figure}

\paragraph{Generated grid of architectures:}
We generate a set of architectures using previously described process by creating a discretized grid of $\xi, \omega, \alpha$ values. Specifically we let $\xi$ be in the interval $[1,16]$, in steps of $1$, to generate 16 layer VGG-D like architectures with different feature maximum locations. We vary the scale $\omega$ in the interval $[0.5, 5.5]$ in steps of $0.5$ and the shape $\alpha$ in the interval $[-40, 40]$ in steps of $4$. That is, we keep the network's functional sequence (including pooling and activation functions and last two fully-connected layers) the same as the original VGG-D architecture and only redistribute the features across different layers. Not all combinations of $\xi, \omega, \alpha$ are considered "valid" as the resulting integral would violate the assumption of keeping the amount of features constant. We thus only take into account combinations that do not lower the total amount of features by more than $5 \%$.  These parameters result in 203 architectures trained on each dataset. 

\paragraph{Training hyper-parameters:}
We train all networks for 150 epochs for CIFAR-10, and 30 epochs for MNIST and Fashion-MNIST using the weight initialization of He et al \cite{He2015}. To make sure that all networks are able to train to convergence, we include batch-normalization with a value of $10^{-4}$ \cite{Ioffe2015} and cycle the learning rate with warm restarts \cite{Loshchilov2017}. We start with an initial learning rate of $10^{-2}$ and continuously lower it to $10^{-5}$ with a restart cycle of 10 epochs, that is then doubled after each restart. To be consistent with evaluation in the literature, we train using a batch size of 128, a weight-decay of $5 \cdot  10^{-4}$ and apply horizontal flip and four pixel random translation data augmentation to the CIFAR-10 data. MNIST and Fashion-MNIST images are resized to $32 \times 32$ to allow for the use of the same architectures. No further data pre-processing or augmentation is applied.
 
\paragraph{Results:}
The validation accuracy for trained architectures parametrized by $\xi, \omega, \alpha$ is shown in figure \ref{fig:results}. We remind the reader that all architectures approximately have the same amount of overall features. Depending on the precise distribution of features the representational capacity can vary. The total amount of parameters is therefore also encoded by marker sizes. Note that for the majority of architectures there is minor variation, with exception of the edge cases where a large amount of features is attributed in the very last two fully-connected layers, where the overall amount of parameters is then smaller. In all three examples, most clearly for CIFAR-10 due to larger accuracy variation, we observe the following trends: 
\begin{itemize}
	\item The accuracy rises with lower $\xi$ value, i.e. architectures that favor larger amounts of features in early layers seem to achieve better accuracy.
	\item The accuracy rises with higher $\omega$ value. This is because low scale values lead to tails of the distribution that map to very little overall amount of features, e.g. only 2 features in a layer. 
\end{itemize} 
\begin{figure}[t]
	\includegraphics[width=\textwidth]{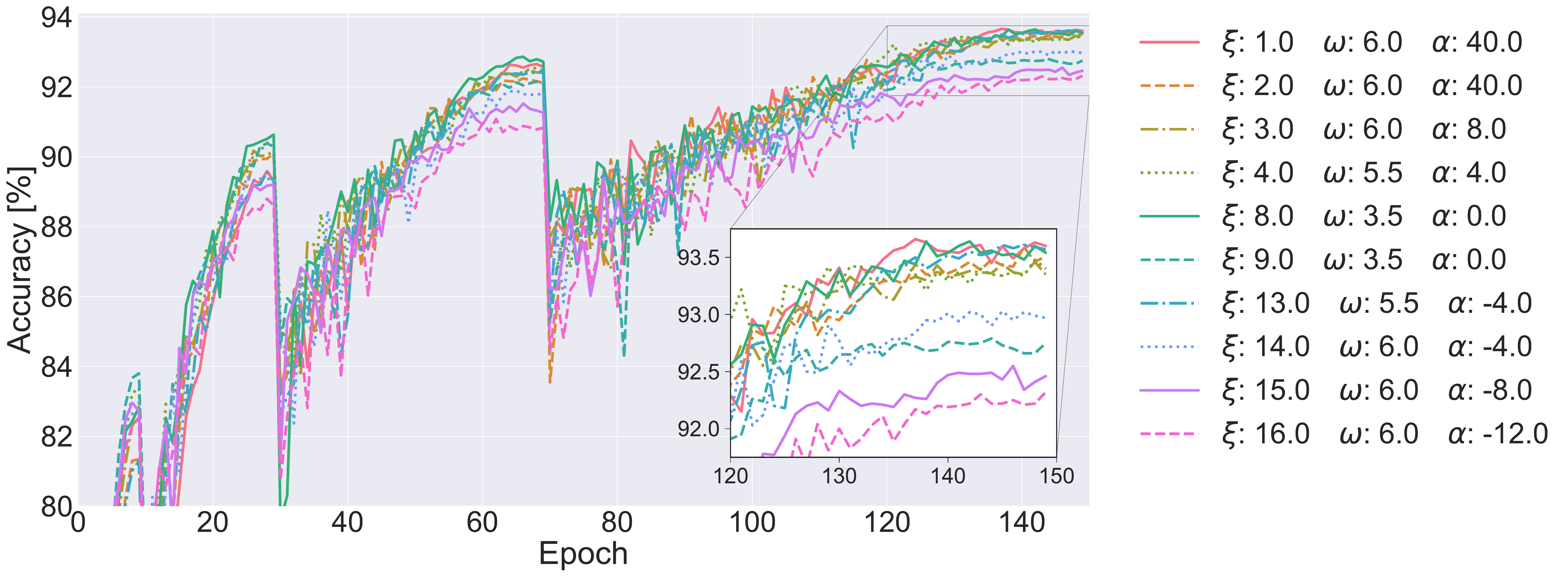}
	\caption{Validation accuracy per epoch of 16-layer VGG-type architectures parametrized through combinations of parameters $\xi, \omega, \alpha$ of a univariate skew normal distribution. The best architectures per $\xi$ are visualized to demonstrate the trend that lower values of $\xi$ are correlated with higher accuracy.}
	\label{fig:training}
\end{figure}

For small or large values of $\xi$, we also observe a constraint of $\alpha$ to a positive or negative range respectively, to make sure that the overall amount of features stays approximately constant. The changing number of total parameters at constant feature amount highlights a different (mal)practice in CNN design, where we generally design amounts of unique features independently of spatial kernel dimensions or questioning the effects on overall parameter count. To emphasize the differences in accuracy, we visualize the best CIFAR-10 architecture per $\xi$ in figure \ref{fig:training}. The tendency of rising accuracy with lower $\xi$ is in contrast with our common assumption of increasing, or even doubling the amount of features as we progress deeper into the CNN layers. We remark that all models converge after 150 epochs. However, the hyper-parameters are selected based on original VGG architectures (i.e. large $\xi$) and not tuned to best fit presented small $\xi$ variants. Additional experiments with 10 VGG-type layers confirm described trends. Due to space constraints we include these results with the open-source code for this work: \url{https://github.com/MrtnMndt/Rethinking_CNN_Layerwise_Feature_Amounts}. 

After analysis of the results presented in figure \ref{fig:training}, we note that the middle range of $\xi$ is difficult to parametrize due to a non constant total number of features for many distribution parameter configurations. In hindsight, one idea could thus be to use a distribution with constant probability mass as the parameters change. One such distribution for further experimentation could be the Beta distribution, with layers binned to equally sized intervals in the $[0,1]$ range.

\section{Conclusion}
We have parametrized CNN architectures with respect to their relative amounts of features per layer using a skew normal distribution. Although further investigation with larger datasets is necessary, our experiments indicate that our historically grown assumption of increasing layer-wise feature counts with increasing network depth is challenged by architectures that favor large early layers. While it isn't emphasized in the original work, architectures generated through the recent trend of reinforcement learning based search seem to be in favor of this trend \cite{Zoph2017}. It will thus be interesting to extend our examination to models with skip connections to see if a similar conclusion hold.
A remaining crucial open question is the reason behind the observed pattern. Is it simply that using too few features in initial layers acts as a bottleneck, making it harder for the remaining layers to retrieve the information about the image that is necessary for classification? Or is there a deeper reason? We motivate to rethink this design principle and more thoroughly analyse future CNN designs. 

\section{Acknowledgements}
\setlength\intextsep{0pt}
\begin{wrapfigure}[2]{l}{0.08\textwidth}
	\includegraphics[width=0.08 \textwidth]{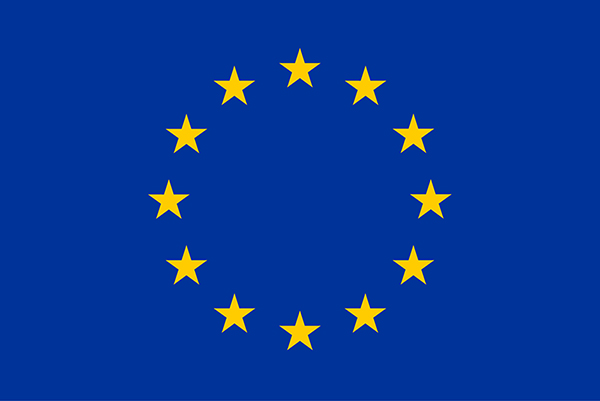}
\end{wrapfigure}
This project has received funding from the European Union's Horizon 2020 research and innovation programme under grant agreement No. 687383 .

\small

\bibliographystyle{unsrtnat}
\bibliography{references}

\end{document}